\pdfoutput=1

\documentclass[11pt]{article}

\usepackage{ACL2023}

\usepackage{times}
\usepackage{latexsym}
\usepackage{xurl}

\usepackage[T1]{fontenc}

\usepackage[utf8]{inputenc}

\usepackage{microtype}

\usepackage{inconsolata}

\usepackage{graphicx}

\title{When a language model is optimized for reasoning, does it still show embers of autoregression? An analysis of OpenAI o1}

\author{R.\ Thomas McCoy,\textsuperscript{1} Shunyu Yao,\textsuperscript{2} Dan Friedman,\textsuperscript{3} Mathew D.\ Hardy,\textsuperscript{4} Thomas L.\ Griffiths\textsuperscript{5,3} \\
  \textsuperscript{1}Department of Linguistics and Wu Tsai Institute, Yale University \hspace{1cm} \textsuperscript{2}OpenAI \\
  \textsuperscript{3}Department of Computer Science, Princeton University \\
  \textsuperscript{4}Roundtable \hspace{1cm} \textsuperscript{5}Department of Psychology, Princeton University \\
  \texttt{tom.mccoy@yale.edu}, \texttt{shunyuy@princeton.edu}, \texttt{dfriedman@princeton.edu}, \\
  \texttt{matt@roundtable.ai}, \texttt{tomg@princeton.edu} \\}

\begin{document}
\maketitle
\begin{abstract}
In ``Embers of Autoregression'' \cite{mccoy2023embers}, we showed that several large language models (LLMs) have some important limitations that are attributable to their origins in next-word prediction.
Here we investigate whether 
these issues
persist with o1, a new system from \mbox{OpenAI} that differs from previous LLMs in that it is optimized for reasoning.
We find that o1 substantially outperforms previous LLMs in many cases, with particularly large improvements on rare variants of common tasks (e.g., forming acronyms from the second letter of each word in a list, rather than the first letter). 
Despite these quantitative improvements, however, o1 still displays the same qualitative trends that we observed in previous systems. Specifically, o1---like previous LLMs---is sensitive to the probability of examples and tasks, performing better and requiring fewer ``thinking tokens'' in high-probability settings than in low-probability ones. 
These results show that optimizing a language model for reasoning can mitigate but might not fully overcome the language model's probability sensitivity.
\end{abstract}

\section{Introduction}

How can we reason about the strengths and limitations of AI systems? In \citet{mccoy2023embers}, we argue that one productive approach is to analyze the system through the lens of the pressures that have shaped it \citep{marr1982vision,shepard1987toward,anderson1990adaptive,griffiths2020understanding}. By considering these pressures, we can make predictions about what strategies the AI system is likely to adopt. Reasoning about these strategies can then provide hypotheses about which types of examples the system will be able to handle well or poorly.

\begin{figure*}[th]
    \centering
    \includegraphics[width=0.65\linewidth]{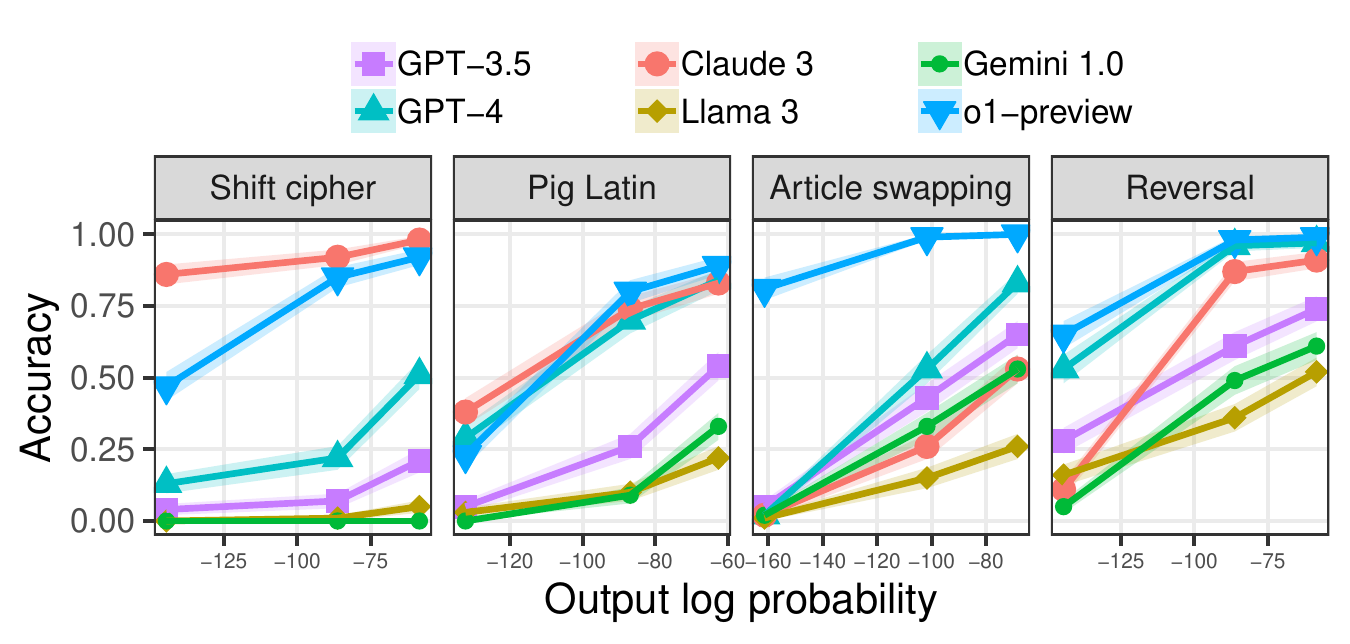}
    \caption{Across the four tasks we considered (shift ciphers, Pig Latin, article swapping, and reversal), all six LLMs evaluated here---including o1---show sensitivity to output probability, with higher accuracies on examples that have a high output probability than on examples that have a low output probability. The results for all models except o1 are from \citet{mccoy2023embers}. The intervals around the lines show one standard error.}
    \label{fig:outputprob}
\end{figure*}

In our prior work, we applied this approach---which we call the \textbf{teleological perspective}---to large language models (LLMs). Perhaps the most significant pressure shaping these systems is their primary training objective of autoregression (next-word prediction; \citealt{elman1990finding,radford2018improving}): they are trained to take in the start of a piece of text and probabilistically predict what word will come next. By considering the probabilistic nature of this objective, we predicted that LLMs would be sensitive to both the probability of the text they need to produce and the commonness of the task they are being asked to perform. These hypotheses were supported by a range of experiments. For example, LLMs performed better at reversing a list of words when the output of the reversal was a high-probability word sequence than when it was a low-probability word sequence.
Thus, even when LLMs are being used for tasks that seem very different from next-word prediction, their performance still shows \textbf{embers of autoregression}---behavioral patterns that result from the influence of being optimized to perform next-word prediction.

In this work, we analyze a new system from OpenAI called o1\footnote{\url{https://openai.com/index/learning-to-reason-with-llms/}} to see whether it also displays these embers of autoregression. Unlike previous LLMs, o1 was explicitly optimized to perform reasoning. Thus, it is possible that this departure from the next-word prediction objective would make o1 less susceptible to the limitations that arise from next-word prediction. On the other hand, it is likely that o1's training involves next-word prediction as well as reasoning optimization, meaning that o1 may still show the effects that arise from next-word prediction.

We find that o1 improves substantially over previous LLMs in many of our evaluations, but it still shows the same qualitative behavioral patterns that we observed with other LLMs. On the front of example probability, o1 scores substantially better on examples with high-probability outputs than ones with low-probability outputs. On the front of task probability, o1 sometimes scores better on common task variants than rare ones, though these task frequency effects are less pronounced than in previous LLMs. In addition to assessments based on accuracy, o1 also provides another way to quantify difficulty, namely via the number of tokens that it produces while working toward an answer. This metric corroborates the results based on accuracy: o1 uses more tokens to produce its answers for low-probability examples and rare task variants than it does for high-probability examples and common task variants. Overall, then, o1 represents an impressive advance on the types of tasks we consider, but it has not fully overcome the issues highlighted in our previous work. 

\section{Background: o1}

The exact details of how o1 works are not publicly available, but a general description of its operation is available at \url{https://openai.com/index/learning-to-reason-with-llms/}. o1 is trained via reinforcement learning to solve reasoning problems using a chain of thought \cite{nye2021show,wei2022chain,kojima2022large}, in which it breaks the problem down into steps before producing the final answer. The user is only given the final answer (not the chain of thought), but the number of tokens inside the chain of thought is provided, so we can tell how long the chain of thought was even though we cannot tell what its contents were; below we have some analyses based on these counts of so-called ``thinking tokens.''

\begin{figure*}[t]
    \centering
    \includegraphics[width=0.65\linewidth]{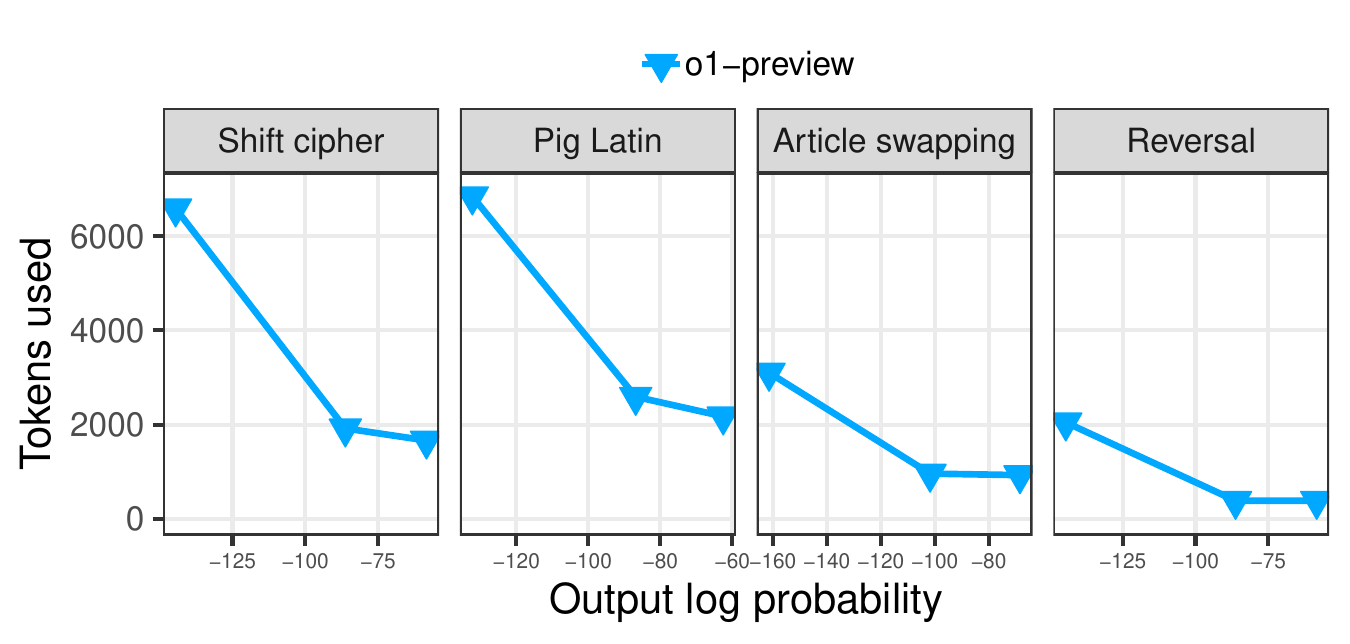}
    \caption{o1 tends to use more tokens when processing examples that have low-probability answers than examples that have high-probability answers. The plots show the median number of tokens that o1 used for each group of examples.}
    \label{fig:outputprobtokens}
\end{figure*}

\section{Results}

The version of o1 used for all results below is \texttt{o1-preview-2024-09-12}, which we tested with its default settings. For detailed descriptions of the tasks and datasets that are evaluated on, see \citet{mccoy2023embers}. We evaluated on only a subset of the tasks from \citet{mccoy2023embers}, excluding those whose datasets involved a large number of examples because o1 has a fairly high cost per example.

\subsection{Output probability}

The first major effect that we tested for was sensitivity to output probability: Does o1 perform better on examples for which the answer is a high-probability string than on examples for which the answer is a low-probability string? We investigated the effects of output probability across four tasks: decoding shift ciphers (a simple type of cipher), decoding messages expressed in Pig Latin, article swapping (swapping certain words in a sequence with the words before them), and reversing a list of words.

As shown in Figure~\ref{fig:outputprob}, o1---like the other LLMs illustrated there---shows clear effects of output probability. For example, in the shift cipher task, its accuracy ranges from 47\% in the lowest-probability case to 92\% in the highest-probability case. Although o1 shows the same qualitative trend as other LLMs, it often outperforms them quantitatively, with particularly strong results in the article swapping task.

In addition to evaluating accuracy, we also noted how many tokens were used by o1 to answer its queries (Figure~\ref{fig:outputprobtokens}). Across all four tasks, o1 tended to use more tokens for low-probability examples than high-probability ones, further corroborating the conclusion that low-probability cases are harder for o1 than high-probability cases.

\subsection{Task frequency}

The other major effect that we tested for was sensitivity to task frequency: Does o1 perform better on task variants that occur frequently in training data (e.g., sorting a list into alphabetical order) than rarer variants of those tasks (e.g., sorting a list into reverse alphabetical order)? For this set of experiments, we considered five task types, with a common and rare variant for each one: decoding messages written in shift ciphers, encoding messages into Pig Latin, forming acronyms, applying a linear function, and sorting a list.

\begin{figure*}
    \centering
    \begin{minipage}{0.7\textwidth}
        \includegraphics[scale=0.45]{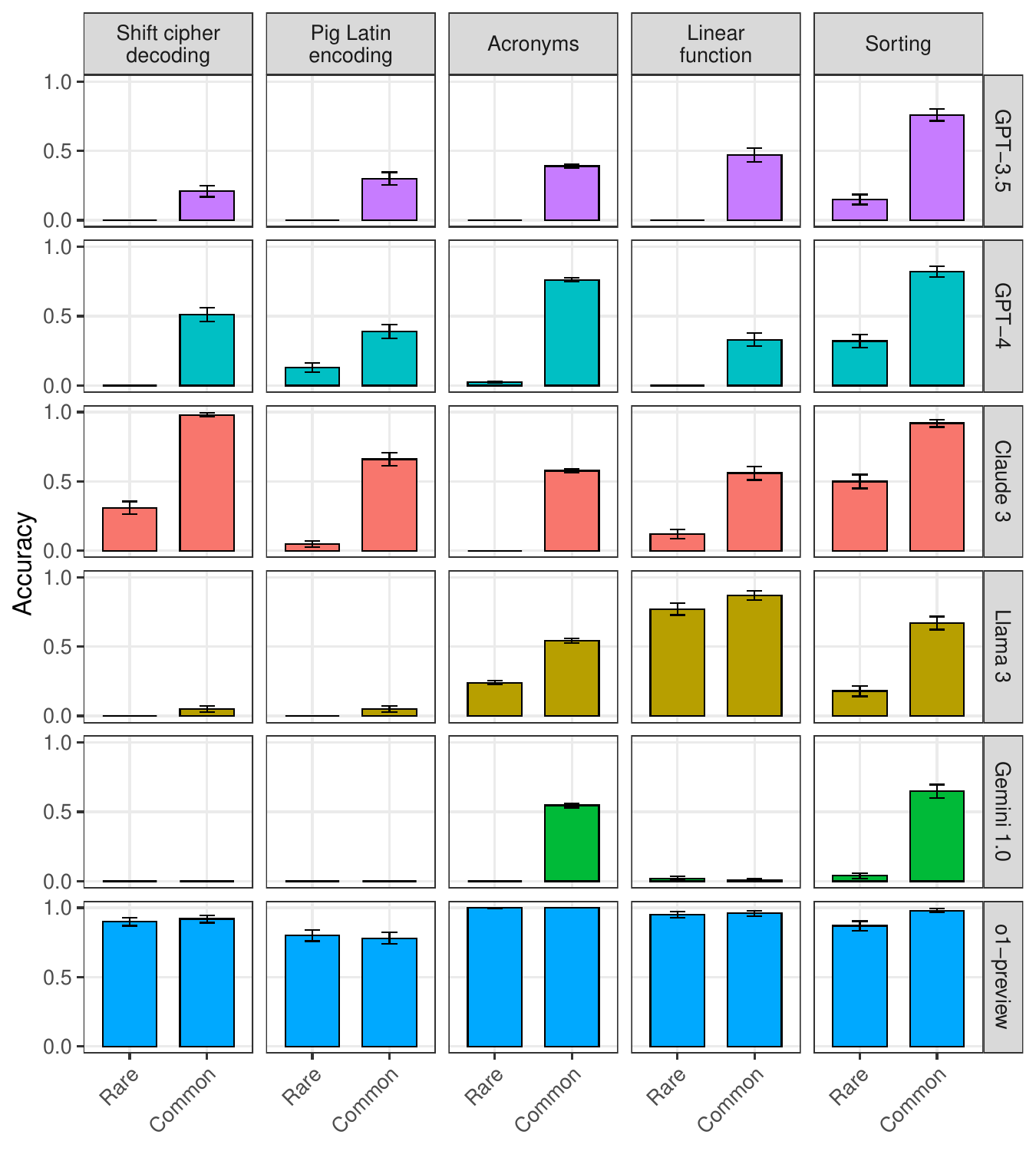}
    \end{minipage}
    \begin{minipage}{0.28\textwidth}
    \centering
        \includegraphics[scale=0.45]{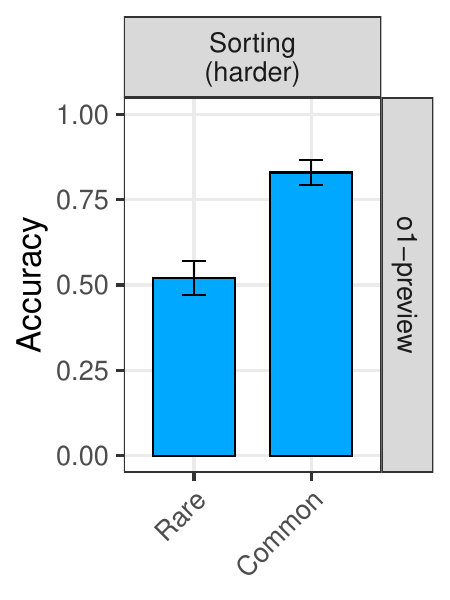}
    \includegraphics[scale=0.45]{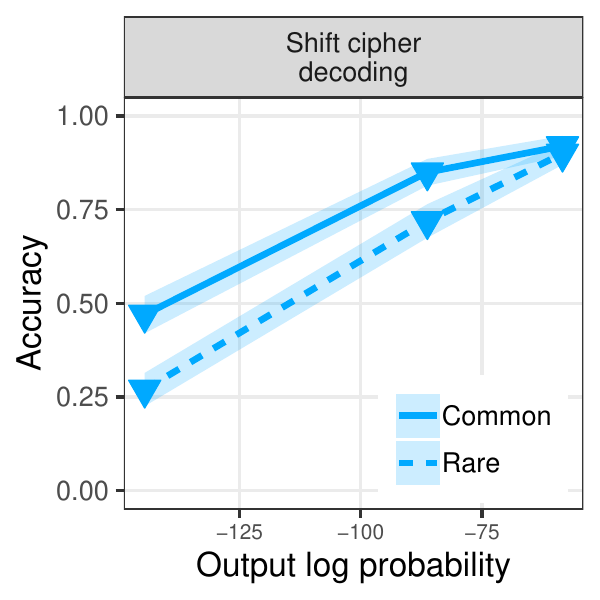}
    \end{minipage}
    \caption{\textbf{Left:} We evaluated LLMs on two variants of five tasks---a variant that is common in Internet text (e.g., forming acronyms from the first letter of each word in a sequence) and a variant that is rare (e.g., forming acronyms from the second letter of each word in a sequence). On these datasets, the five LLMs other than o1 showed much higher accuracy on the common variants than the rare ones, but o1 showed similar performance on common and rare variants. The results for models other than o1 are from \citet{mccoy2023embers}. \textbf{Top right:} On datasets based on challenging sorting tasks, o1 performs better on the common type of sorting (i.e., sorting into alphabetical order) than on the rare type of sorting (i.e., sorting into reverse alphabetical order). \textbf{Bottom right:} When decoding shift ciphers, o1 shows roughly the same performance on the common cipher type and on the rare cipher type when the examples are ones with a high output probability. However, when it is instead evaluated on examples with medium or low probability, its accuracy is higher for the common cipher type than the rare one. The error intervals in all plots show one standard error.}
    \label{fig:taskfreq}
\end{figure*}

We find that o1 performs substantially better than the other LLMs on the rare task variants (Figure~\ref{fig:taskfreq}, left). Further, although all other LLMs show stark differences between the rare and common versions of at least some tasks, o1 achieves similar scores between the two members of each pair. These results suggest that o1 might not be sensitive to task frequency, but it is difficult to draw definitive conclusions because o1's strong performance might be producing ceiling effects. That is, even if o1 is sensitive to task frequency, the datasets used here might not be challenging enough for the effects to be evidenced.

To address the possibility of ceiling effects, we investigated more challenging versions of two of the tasks. First, the sorting tasks involve sorting a list of words into alphabetical order (the common variant) or reverse alphabetical order (the rare variant). We made sorting more challenging by having all words in the list start with the same letter---namely, \textit{i}---so that finding the right ordering requires considering at least the first two letters of each word, whereas previously it was usually sufficient to only consider the first letter. In this harder version of sorting, o1 now performs substantially better on the common version of the task than the rare one (Figure~\ref{fig:taskfreq}, top right). Second, the shift cipher tasks involve decoding a message written in a simple cipher, where the cipher involves shifting each letter forward in the alphabet either 13 positions (the common variant) or 12 positions (the rare variant). To modulate difficulty in this case, we used examples whose target outputs varied in probability, since we have established that lower-probability cases tend to be harder for o1. Although o1 performs similarly on the common and rare task variants in the highest-probability case, its performance in the medium-probability and low-probability settings is higher for the common task variant than the rare one (Figure~\ref{fig:taskfreq}, bottom right). These additional experiments therefore show that o1 is sensitive to task frequency in at least some cases---but this trend may only be observable when the examples are challenging enough for bring o1's performance substantially below 100\% accuracy.

\begin{figure*}
    \centering
    \includegraphics[width=\linewidth]{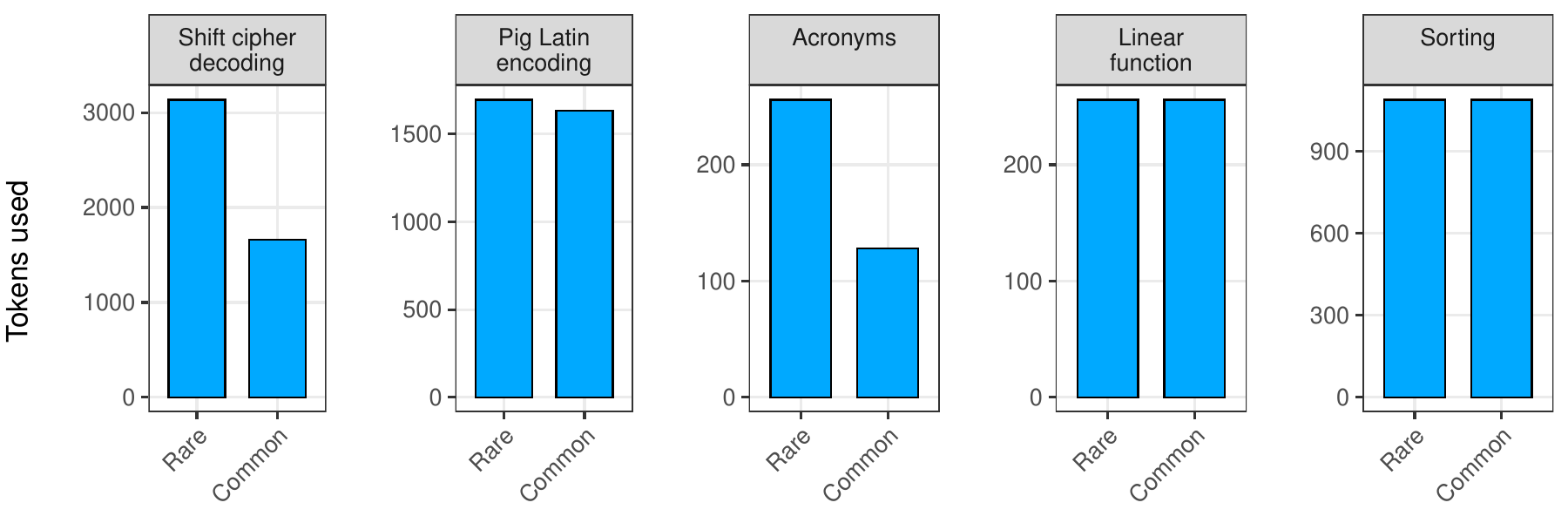}
    \caption{In some cases---namely, for shift ciphers and acronyms---o1 consumes more tokens when performing a rare task variant than a common task variant. For the other task pairs, the number of tokens it consumes is similar across both task frequency levels. The bars show the median number of tokens used within each group of examples. Note that the vertical axes have different scales in each plot.}
    \label{fig:pairs_tokens}
\end{figure*}

Finally, plotting the number of tokens that o1 uses for each task variant reveals additional evidence that rare task variants can be harder for o1 than common task variants (Figure~\ref{fig:pairs_tokens}). Specifically, for both shift cipher decoding and acronyms, o1 uses far more tokens for the rare task variant than the common one. Notably, for both of these tasks, accuracy is nearly identical for both task variants; e.g., for acronyms, o1 achieved 100\% accuracy on the common variant and 99.9\% accuracy on the rare variant. These cases therefore show that it is possible for o1 to display a difference in difficulty as quantified by the number of tokens that are used even when the relevant accuracies show no variation. Although shift cipher decoding and acronyms both showed large differences in token quantities between the two task variants, the other three task types had almost identical token usage between variants, showing that differences in task frequency are only sometimes associated with differences in token usage.

Overall, o1 shows substantially less sensitivity to task frequency than the other LLMs we have previously identified. However, there still is evidence of task frequency effects in some cases, namely when the tasks are made more challenging and when we consider the number of tokens consumed by o1. We therefore conclude that o1 can be substantially influenced by task frequency.

\section{Conclusion}

On many of the tasks we considered, o1 performed substantially better than the LLMs we had previously evaluated, with particularly strong results on rare variants of common tasks. However, it still qualitatively showed both of the central types of probability sensitivity discussed in \citet{mccoy2023embers}: sensitivity to output probability and sensitivity to task frequency.

These results are consistent with the teleological perspective that we have argued for. On one hand, o1 is explicitly optimized for reasoning, so we should expect it to perform well on the sorts of algorithmic tasks that we have considered---as it indeed does. On the other hand, although this is not explicitly stated in the o1 documentation as far as we can tell, o1 also probably went through a substantial amount of training on next-word prediction, such that we would expect it to display the behavioral signatures that go with being optimized for next-word prediction---and we have indeed found that it does so. These results support the view that developing a complete teleological analysis of an AI system requires consideration of all types of optimization that have been applied to that system.

We see two potential aspects of o1 that might give rise to the probability sensitivity we have observed. First, probability sensitivity might arise during the process of generating text, for the same reasons as it does in other types of LLMs---the generation process in any system optimized for statistical prediction is expected to be biased toward high-probability text. Indeed, \citet{prabhakar2024deciphering} showed that LLMs using chain-of-thought reasoning are susceptible to probability effects when generating text, so it would not be surprising if the process of generation produces similar effects in the hidden chains of thought produced by o1. Second, it might be that the process of developing a chain of thought could also introduce biases toward high-probability scenarios: if o1's task is viewed as considering multiple potential chains of thought and deciding between them, this decision might be at least partially influenced by probability (e.g., favoring chains that produce higher-probability answers because those answers are judged as more plausible), which would introduce biases favoring high-probability text (or would enhance those biases if they are already present).

It is not clear what modeling enhancements would suffice to fully overcome the limitations that we have highlighted. One potential solution would be to incorporate model components that do not involve probabilistic judgments in any way, such as modules that execute Python code. For now at least, the sparks of AGI \citep{bubeck2023sparks} that LLMs may be producing continue to be accompanied by embers of autoregression.

\section*{Competing interests}

S.Y.\ is employed by OpenAI, but this work is an addendum to a project that was completed before he started at OpenAI. Though this paper includes some speculation about how o1 works, S.Y.\ did not contribute to these parts of the paper, so the paper should not be viewed as providing any information about how o1 works beyond what is publicly available.

\bibliography{custom}
\bibliographystyle{acl_natbib}

\end{document}